\pdfoutput=1

\documentclass[11pt]{article}

\usepackage[final]{acl}

\usepackage{times}
\usepackage{latexsym}

\usepackage[T1]{fontenc}

\usepackage[utf8]{inputenc}

\usepackage{microtype}

\usepackage{inconsolata}

\usepackage{graphicx}
\usepackage{ragged2e}

\usepackage{amsfonts}
\usepackage{booktabs}
\usepackage{multirow}
\usepackage{threeparttable}

\title{Representation-based Broad Hallucination Detectors Fail to Generalize Out of Distribution}

\author{
  \textbf{Zuzanna Dubanowska\textsuperscript{1}},
  \textbf{Maciej \.Zelaszczyk\textsuperscript{1}},
  \textbf{Micha\l{} Brzozowski\textsuperscript{1}},
  \textbf{Paolo Mandica\textsuperscript{1}},
\\
  \textbf{Micha\l{} Karpowicz\textsuperscript{1}}
\\
  \textsuperscript{1}Samsung AI Research Center, Warsaw, Poland
\\
  \small{
    \textbf{Correspondence:} \href{mailto:z.dubanowska@samsung.com}{z.dubanowska@samsung.com}
  }
}

\begin{document}
\maketitle
\begin{abstract}
We critically assess the efficacy of the current SOTA in hallucination detection and find that its performance on the \texttt{RAGTruth} dataset is largely driven by a spurious correlation with data. Controlling for this effect, state-of-the-art performs no better than supervised linear probes, while requiring extensive hyperparameter tuning across datasets. Out-of-distribution generalization is currently out of reach, with all of the analyzed methods performing close to random. We propose a set of guidelines for hallucination detection and its evaluation.
\end{abstract}

\section{Introduction}
While LLMs \cite{grattafiori2024llama3herdmodels, jiang2023mistral7b, bai2023qwentechnicalreport, biderman2023pythia} have made significant progress in various scenarios, they still display undesirable behavior, known as \textit{hallucinations}, which undermines their reliability in both critical and everyday tasks. Hallucinations have been extensively studied, with a wide range of detection methodologies proposed depending on the level of access to model internals. \textbf{Black-box methods} operate on model input-output behavior. These include frameworks verifying generated responses against context \cite{es2023ragasautomatedevaluationretrieval, saadfalcon2024aresautomatedevaluationframework, hu2024refcheckerreferencebasedfinegrainedhallucination}, evaluating self-consistency across multiple response variants \cite{ manakul2023selfcheckgptzeroresourceblackboxhallucination} or multiple verification steps \cite{friel2023chainpollhighefficacymethod}, querying the model to evaluate the truthfulness of it's own response \cite{kadavath2022languagemodelsmostlyknow} or training classifiers to flag inconsistencies \cite{kovács2025lettucedetecthallucinationdetectionframework}. \textbf{Grey-box methods} center on uncertainty quantification through metrics like token-level log probabilities \cite{hu-etal-2024-embedding} or entropy of token distribution \cite{farquhar2024detecting}. Recent \textbf{white-box methodologies} \cite{du2024haloscope, sriramanan2024llm} have focused on detecting hallucinations by leveraging internal representations, moving beyond text-based detection. A line of work learns to predict the hallucination by training linear probes on the hidden states of LLMs directly \cite{azaria2023internalstatellmknows, li2024inferencetimeinterventionelicitingtruthful} or to approximate derived metrics like semantic entropy \cite{kossen2024semanticentropyprobesrobust} or semantic consistency \cite{chen2024inside}. Among white-box methods, many recent approaches rely on analyzing the attention mechanisms of Transformer models, which has garnered significant research interest. Working under the assumption that \textit{retrieval heads} \cite{wu2024retrievalheadmechanisticallyexplains} are an important mechanism for information propagation within LLMs, \citet{gema2024decoredecodingcontrastingretrieval} show that masking the retrieval heads leads to increased hallucination incidence and that contrasting the next-token predictions of the base model with those of the masked one can act as a mitigation mechanism. \citet{sun2025redeep} further explores this attention-based methodology, which we analyze in depth in Section \ref{sec:sota}. Treating all types of hallucinations as a single category is challenging, resulting in methods that perform well in some scenarios (e.g., hallucination detection in summarization tasks) but poorly in others. Retrieval-Augmented Generation (RAG) has gained traction as a method to potentially decrease hallucinations. Nevertheless, even with RAG, models can still fail to correctly attend to the context or overwrite it with parametric knowledge. Due to the limited availability of human-labeled RAG hallucination datasets, studying hallucinations related to contextual errors is difficult, with only a few attempts so far \cite{sun2025redeep, ravi2024lynxopensourcehallucination, kovács2025lettucedetecthallucinationdetectionframework}.
We focus on the current SOTA approaches for detecting hallucinations using internal representations, with the goal of defining a clearer future trajectory for hallucination detection. Our contributions are as follows. \textbf{(A)} We verify hallucination detectors based on model internals via linear probes, a random forest classifier and sparse-autoencoder (SAE) probes. \textbf{(B)} Assess SOTA and find that most of its performance is due to spurious correlation rather than genuine hallucinatory signal. \textbf{(C)} Analyze detection out-of-distribution and find that generalization remains a challenge. \textbf{(D)} Find that SAE features do not provide consistent benefits for hallucination detection.

\section{State of the art}
\label{sec:sota}


\texttt{ReDeEP} \cite{sun2025redeep} employs internal model mechanisms to identify hallucinations, specifically those cases where external context serves as the ground truth but may be compromised by the model's parametric knowledge. It relies on copying heads \cite{elhage2021mathematical} and assumes that attention scores are an appropriate proxy for what information the model integrates from the context. Moreover, the parametric interventions of the model, measured by the Jensen-Shannon divergence between pre- and post-MLP representations, serve as an important hallucination indicator. They hypothesize that increased attention scores correlate with a decrease in the hallucination incidence, while elevated parametric interventions correspond to increased occurrence of hallucinations. \texttt{ReDeEP} comes in two flavors: token- and chunk-based, where the attention and parametric scores are measured on individual tokens or token chunks, respectively. \texttt{ReDeEP} achieves SOTA hallucination detection results on two publicly available RAG hallucinations datasets: \texttt{RAGTruth} and \texttt{Dolly (AC)}. 

\citet{azaria2023internalstatellmknows} introduce \texttt{SAPLMA}, another representation-based hallucination detection method. They investigate whether the LLM possesses some notion of truthfulness of generated statements in its hidden state. The authors postulate that if it does, this information can be used to detect and mitigate hallucinations. Based on experimental results, they hypothesize that the hidden states indeed encode signals of truth or falsehood, leading to the hypothesis that the model may "know" when it hallucinates. \texttt{SAPLMA} performs on-par with \texttt{ReDeEP} in hallucination detection on \texttt{RAGTruth}.

\section{Spurious correlation explains SOTA performance}
\label{sec:naive}

\texttt{ReDeEP}'s evaluation focuses exclusively on the performance across all three subtasks in the \texttt{RAGTruth} dataset. This prompts us to investigate its performance on individual tasks. Table \ref{tab:redeep_per_task_results} reveals significant variability in performance across individual tasks, with overall performance generally lower than that on the entire dataset. To understand the disparity in performance between tasks, we examine the composition of the \texttt{RAGTruth} dataset (details in Appendix \ref{app:ragtruth}).

\begin{table}[ht]
  \centering
  \resizebox{\columnwidth}{!}{
  \begin{tabular}{llccccc}
    \toprule
    \textsc{Model} & \textsc{Task Type} & \textsc{AUC} & \textsc{PCC} & \textsc{Precision} & \textsc{Recall} & \textsc{F1} \\
    \midrule
    LLaMA-2 7B Chat & D2T & 0.3951 & -0.1246 & \textbf{0.7931} & \textbf{0.748} & \textbf{0.7699} \\
                    & QA & \textbf{0.6360} & \textbf{0.2122} & 0.4528 & 0.4615 & 0.4571 \\
                    & Summary & {0.5767} & {0.1121} & {0.4839} & 0.2941 & 0.3659 \\
                    & Overall & {0.7324} & 0.3978 & {0.7217} & 0.6770 & 0.6986 \\
    \midrule
    LLaMA-2 13B Chat & D2T & \textbf{0.6679} & {0.1545} & \textbf{0.9357} & \textbf{0.9493} & \textbf{0.9424} \\
                     & QA & {0.6503} & \textbf{0.2259} & 0.3103 & 0.7500 & 0.4390 \\
                     & Summary & 0.5342 & 0.0424 & 0.2500 & 0.7879 & 0.3796 \\
                     & Overall & {0.8177} & {0.5484} & 0.5875 & {0.8599} & 0.6980 \\
    \midrule
    LLaMA-3 8B Instruct & D2T & 0.5084 & -0.0072 &\textbf{ 0.8707} & 0.7652 &\textbf{ 0.8145} \\
                        & QA & \textbf{0.5974} & \textbf{0.1573} & {0.5514} & {0.7564} & {0.6378} \\
                        & Summary & 0.5593 & 0.1102 & 0.2475 & \textbf{0.8065} & 0.3788 \\
                        & Overall & 0.7534 & 0.4512 & 0.6465 & 0.8000 & {0.7151} \\
    \bottomrule
  \end{tabular}
  }
  \caption{ReDeEP (token) performance on RAGTruth per task type. Best task results (excluding \texttt{Overall}) are bolded.}
  \label{tab:redeep_per_task_results}
\end{table}

\begin{table}[ht]
\centering
\resizebox{\columnwidth}{!}{
\begin{tabular}{lcc}
\toprule
\textsc{Model} & \textsc{Task Type} & \textsc{Hallucination Rate} \\
\midrule
\multirow{3}{*}{LLaMA-2 7B} 
  & D2T & 0.8596 \\
  & QA & 0.5157 \\
  & Summary & 0.4602 \\
\midrule
\multirow{3}{*}{LLaMA-3 8B Instruct} 
  & D2T & 0.8800 \\
  & QA & 0.5200 \\
  & Summary & 0.2200 \\
\bottomrule
\end{tabular}
}
\caption{Hallucination rates per task type on LLaMA-2 7B and LLaMA-3 8B. See Appendix \ref{app:additional}, Table \ref{tab:hallucination_all_models} for other models.}
\label{tab:hallucination_rates}
\end{table}

The fraction of model responses which human annotators labeled as hallucinations (hallucination rate) is notably higher for the \texttt{D2T} task (Table \ref{tab:hallucination_rates} and Table \ref{tab:hallucination_all_models}). This means that a random sample from the \texttt{D2T} task is far more likely to be hallucinatory than for other tasks, leading to a strong correlation between task type and the hallucination label, which may be exploited by \texttt{ReDeEP}. 


We construct a \textit{na\"ive classifier} which follows a heuristic: it predicts 1 (i.e., hallucinated sample) if the input's task type is \texttt{D2T}, and 0 for all other tasks. In Table \ref{tab:model_performance}, we compare the results of the na\"ive classifier with those reported in the \texttt{ReDeEP} paper for the \texttt{RAGTruth} test set. The AUC metrics are very similar, and the na\"ive classifier even outperforms \texttt{ReDeEP} in terms of the Pearson correlation between the true hallucination label and the model score. This suggests that the high scores on the overall \texttt{RAGTruth} benchmark are mostly due to the imbalance in the hallucination incidence between the tasks. Additionally, our na\"ive classifier can outperform \texttt{SEP} \cite{kossen2024semanticentropyprobesrobust} and \texttt{SAPLMA} \cite{azaria2023internalstatellmknows} - supervised methods.

\begin{table}[ht]
  \centering
  \resizebox{0.5\columnwidth}{!}{
  \begin{tabular}{lcc}
    \toprule
    \textsc{Classifier} & \textsc{AUC} & \textsc{PCC} \\
    \midrule
    na\"ive & 0.7119 & \textbf{0.4494} \\
    SEP & 0.7143 & 0.3355 \\
    SAPLMA & 0.7037 & 0.3188 \\
    ReDeEP (token) & 0.7325 & 0.3979 \\
    ReDeEP (chunk) & \textbf{0.7458} & 0.4203 \\
    \bottomrule
  \end{tabular}
  }
  \caption{Model performance metrics on \texttt{RAGTruth} using LLaMA-2 7B.}
  \label{tab:model_performance}
\end{table}

\begin{table}[ht]
  \centering
  \resizebox{\columnwidth}{!}{
  \begin{threeparttable}
  \begin{tabular}{llcccc}
    \toprule
    \textsc{Eval Dataset} & \textsc{Hyper-params} & \textsc{AUC} & \textsc{PCC} & \textsc{Recall} & \textsc{F1} \\
    \midrule
    RAGTruth & RAGTruth & 0.7541 & 0.4522 & 0.8008 & 0.7148 \\
             & Dolly    & 0.7230 & 0.4413 & 0.6390 & 0.6844 \\
    Dolly    & Dolly    & 0.6223 & 0.2129 & 0.8684 & 0.5841 \\
             & RAGTruth & 0.5005 & -0.0188 & 0.8684 & 0.5410 \\
    \bottomrule
  \end{tabular}
  \begin{tablenotes}
  \item[] The \textsc{Eval Dataset} column specifies the dataset on which the method is evaluated, whereas the \textsc{Hyper-params} column indicates the dataset used to fine-tune the hyperparameters of the method.
  \end{tablenotes}
  \end{threeparttable}
  }
  \caption{Cross-dataset evaluation results of ReDeEP – LLaMA-3 8B.}
  \label{tab:cross-dataset-eval}
\end{table}

\begin{table*}[ht]
  \centering
  \resizebox{\textwidth}{!}{%
  \begin{threeparttable}
  \begin{tabular}{l|cccccccccccc|cccc}
    \toprule
    & \multicolumn{4}{c}{\textsc{Question Answering}} & \multicolumn{4}{c}{\textsc{Data-to-text writing}} & \multicolumn{4}{c}{\textsc{Summarization}} & \multicolumn{4}{c}{\textsc{Overall}} \\
    \cmidrule(r){2-5} \cmidrule(r){6-9} \cmidrule(r){10-13} \cmidrule(r){14-17}
    Method & AUC & Precision & Recall & F1 & AUC & Precision & Recall & F1 & AUC & Precision & Recall & F1 & AUC & Precision & Recall & F1 \\
    \midrule
    ReDeEP\tnote{a} & 0.6360 & 0.4528 & 0.4615 & 0.4571 & 0.3951 & 0.7931 & 0.7480 & \textbf{0.7699} & 0.5767 & 0.4839 & 0.2941 & 0.3659 & 0.7324 & 0.7217 & 0.677 & 0.6986 \\
        Logistic Regression\tnote{b} 
    & $0.6900_{\pm0.02}$ & $\mathbf{0.6912}_{\pm0.01}$ & $0.6900_{\pm0.03}$ & $\mathbf{0.6901}_{\pm0.01}$ 
    & $\mathbf{0.6555}_{\pm0.00}$ & $0.6555_{\pm0.00}$ & $0.7611_{\pm0.00}$ & $0.6777_{\pm0.00}$ 
    & $0.6376_{\pm0.01}$ & $0.6380_{\pm0.02}$ & $0.6376_{\pm0.04}$ & $\mathbf{0.6378}_{\pm0.01}$ 
    & $\mathbf{0.7951}_{\pm0.01}$ & $\mathbf{0.7951}_{\pm0.01}$ & $0.7930_{\pm0.01}$ & $\mathbf{0.7934}_{\pm0.01}$ \\
    Random Forest\tnote{c} 
    & $0.6821_{\pm0.01}$ & $0.6864_{\pm0.01}$ & $0.6821_{\pm0.01}$ & $0.6817_{\pm0.01}$ 
    & $0.5227_{\pm0.00}$ & $0.5227_{\pm0.00}$ & $0.9318_{\pm0.00}$ & $0.5068_{\pm0.00}$ 
    & $\mathbf{0.6410}_{\pm0.02}$ & $\mathbf{0.6598}_{\pm0.02}$ & $\mathbf{0.6410}_{\pm0.01}$ & $0.6376_{\pm0.02}$ 
    & $0.6994_{\pm0.03}$ & $0.6994_{\pm0.02}$ & $0.7191_{\pm0.03}$ & $0.7050_{\pm0.03}$ \\
    SAE Classifier\tnote{d} & $\mathbf{0.7106_{\pm0.00}}$ & $0.5325_{\pm0.00}$ & $\mathbf{0.7885_{\pm0.00}}$ & $0.6804_{\pm0.00}$ & $0.6391_{\pm0.00}$ & $\mathbf{0.8750_{\pm0.00}}$ & $0.7967_{\pm0.00}$ & $0.6170_{\pm0.00}$ & $0.6182_{\pm0.00}$ & $0.4821_{\pm0.00}$ & $0.5294_{\pm0.00}$ & $0.6150_{\pm0.00}$ & $0.7105_{\pm0.00}$ & $0.6655_{\pm0.00}$ & $\mathbf{0.8540_{\pm0.00}}$ & $0.7048_{\pm0.00}$ \\
    SAPLMA\tnote{e} 
    & $0.5699_{\pm0.02}$ & $0.3905_{\pm0.02}$ & $0.500_{\pm0.18}$ & $0.5178_{\pm0.02}$ 
    & $0.5476_{\pm0.01}$ & $0.8200_{\pm0.00}$ & $\mathbf{1.0000_{\pm0.00}}$ & $0.4505_{\pm0.00}$ 
    & $0.5959_{\pm0.03}$ & $0.3976_{\pm0.03}$ & $0.5294_{\pm0.03}$ & $0.5427_{\pm0.03}$ 
    & $0.7486_{\pm0.00}$ & $0.6300_{\pm0.03}$ & $0.7788_{\pm0.05}$ & $0.6479_{\pm0.03}$ \\

    \bottomrule
  \end{tabular}
  \begin{tablenotes}
  \item[a] ReDeEP is a deterministic method. Results do not vary between runs.
  \item[b] Best result - based on layer 15's activations.
  \item[c] Best result - based on layer 5's activations.
  \item[d] \texttt{QA}: layer 15, max activation. \texttt{D2T}: layer 12, last token of response, contrastive. \texttt{Summary}: layer 13, last token of response, contrastive. \texttt{Overall}: layer 13, max activation.
  \item[e] Best result - based on layer 16's activations.
  \end{tablenotes}
  \end{threeparttable}
  }
  \caption{\texttt{RAGTruth} evaluation results - LLaMA-2 7B.}
  \label{tab:ragtruth-eval-per-task-llama2-7b}
\end{table*}

The \texttt{D2T} task is highly specific as it relies on prompts in the JSON format (Appendix \ref{app:ragtruth}, Table \ref{tab:ragtruth}). Therefore, detectors trained and tested on the entire \texttt{RAGTruth} dataset may in reality respond to the presence of JSON in the prompt. To verify that detecting the type of the task from model activations is possible, we collect model activations from the last token and the last layer of the \texttt{Llama2-7B-Chat} model and we use them to train a logistic regression to predict the JSON task. The trained linear probe achieves perfect prediction on the test set, with an AUC of 1.0.

The evaluation problems are not an issue with the \texttt{RAGTruth} dataset itself. In fact, the results in the dataset paper \cite{niu2024ragtruth} are reported per task. This distinction also exists in methods proposed in \cite{song-etal-2024-rag, belyi-etal-2025-luna, kovács2025lettucedetecthallucinationdetectionframework}. However, aggregated metrics have also been reported in \cite{ravi2024lynxopensourcehallucination, sriramanan2024llm}.

\section{Hallucination detection with model internals}
\label{sec:classifier}

The fundamental question we should address first is this: \textit{Is it possible to classify a response from an LLM as hallucinatory based on its internal states?} If this is possible, then it would be a natural benchmark against other methods based on model internals. To provide an answer, we extract the activations of the LLM in the prompt processing and answer generation phases and then use those activations as input to a classifier. Extraction is performed from the residual stream: pre-attention (\texttt{resid\_pre}) and pre-MLP (\texttt{resid\_mid}). We consider linear and non-linear probes (raw activations, details in Appendix \ref{app:activation-probes}), and SAE probes (SAE features, details in Appendix \ref{app:sae-probes}).

\begin{table*}[t]
  \centering
  \resizebox{\textwidth}{!}{
  \begin{threeparttable}
  \begin{tabular}{l|cccccccccccc|cccc}
    \toprule
    & \multicolumn{4}{c}{\textsc{Question Answering}} & \multicolumn{4}{c}{\textsc{Data-to-text writing}} & \multicolumn{4}{c}{\textsc{Summarization}} & \multicolumn{4}{c}{\textsc{Overall}} \\
    \cmidrule(r){2-5} \cmidrule(r){6-9} \cmidrule(r){10-13} \cmidrule(r){14-17}
    Method & AUC & Precision & Recall & F1 & AUC &Precision & Recall & F1 & AUC & Precision & Recall & F1 & AUC & Precision & Recall & F1 \\
    \midrule
    ReDeEP & - & - & - & - & - & - & - & - & - & - & - & - &  0.5163 & 0.4878 & 0.7092 & \textbf{0.578} \\
    Logistic Regression & \textbf{0.5086} & \textbf{0.5000} & 0.5100 & 0.3100 & 0.5072 & 0.5000 & 0.5100 & 0.2300 & 0.5424 & \textbf{0.5100} & 0.5300 & 0.2800 & 0.5139 & \textbf{0.5100} & 0.5100 & 0.3300 \\
    Random Forest & 0.5077 & \textbf{0.5000} & 0.5100 & \textbf{0.4800} & 0.5178 & 0.5100 & 0.5200 & 0.4300 & 0.5049 & 0.5000 & 0.5000 & 0.3400 & 0.5033 & 0.5000 & 0.5000 & 0.4500 \\
    SAE classifier & 0.4889 & 0.4915 & \textbf{0.8112} & 0.4291 & \textbf{0.5273} & \textbf{0.5148} & 0.8531 & \textbf{0.4705} & 0.4856 & 0.4901 & \textbf{0.8671} & 0.3974 & 0.4845 & 0.4841 & 0.5315 & 0.4833 \\
    SAPLMA & 0.5000 & 0.0833 & 0.0051 & 0.4256 & 0.4388 & 0.2603 & \textbf{1.000} & 0.2084 & \textbf{0.5583} & 0.3671 & 0.5686 & \textbf{0.5113} & \textbf{0.5520} & 0.2610 & \textbf{1.000} & 0.2124 \\
    \bottomrule
  \end{tabular}
  \begin{tablenotes}
      \item[] Results presented for layer / hyperparameter combinations performing best on \textsc{RAGTruth} on the \textsc{Overall} task (Table \ref{tab:ragtruth-eval-per-task-llama2-7b}), except for the SAE classifier for which the activation on the last token of the response has been used instead of the max activation.
  \end{tablenotes}
  \end{threeparttable}
  }
  \caption{Evaluation results | train: \texttt{RAGTruth} | eval: \texttt{SQuAD} - LLaMA-2 7B.}
  \label{tab:inter-eval-inter-dataset-train-ragtruth-llama2-7b}
\end{table*}

\begin{table*}[ht]
  \centering
  \resizebox{\textwidth}{!}{
  \begin{threeparttable}
  \begin{tabular}{l|c|cccccccccccc}
    \toprule
    & & \multicolumn{4}{c}{\textsc{Question Answering}} & \multicolumn{4}{c}{\textsc{Data-to-text writing}} & \multicolumn{4}{c}{\textsc{Summarization}} \\
    \cmidrule(r){3-6} \cmidrule(r){7-10} \cmidrule(r){11-14} 
    Method & Eval task & AUC & Precision & Recall & F1 & AUC & Precision & Recall & F1 & AUC & Precision & Recall & F1 \\
    \midrule
    \multirow{3}{*}{Logistic Regression} 
    & \textsc{QA}   & 0.5720 & 0.5700 & 0.5700 & 0.5700 & 0.5596 & 0.5500 & 0.5600 & 0.5500 & 0.5280 & 0.5300 & 0.5300 & 0.5200 \\
    & \textsc{D2T}  & 0.5140 & 0.5600 & 0.5100 & 0.4000 & 0.5564 & 0.6400 & 0.5600 & 0.5600 & 0.5113 & 0.5200 & 0.5100 & 0.4400 \\
    & \textsc{Summ.}& 0.5332 & 0.5400 & 0.5200 & 0.4800 & 0.4464 & 0.4700 & 0.4500 & 0.3900 & 0.6006 & 0.6000 & 0.6000 & 0.6000 \\
    \midrule
    \multirow{3}{*}{Random Forest} 
    & \textsc{QA}   & 0.5886 & 0.5900 & 0.5800 & 0.5900 & 0.5402 & 0.5200 & 0.5400 & 0.5000 & 0.5006 & 0.5000 & 0.5000 & 0.5000 \\
    & \textsc{D2T}  & 0.4910 & 0.4800 & 0.4900 & 0.4300 & 0.5000 & 0.4000 & 0.5000 & 0.4400 & 0.5080 & 0.5200 & 0.5100 & 0.4100 \\
    & \textsc{Summ.}& 0.5115 & 0.5100 & 0.5100 & 0.4900 & 0.5055 & 0.5000 & 0.5100 & 0.3700 & 0.5177 & 0.5300 & 0.5200 & 0.4900 \\
    \midrule
    \multirow{3}{*}{SAE Classifier} 
    & \textsc{QA}   & 0.7055 & 0.5256 & 0.7885 & 0.6742 & 0.5000 & 0.3467 & 1.000 & 0.2574 & 0.5051 & 0.3490 & 1.000 & 0.2688 \\
    & \textsc{D2T}  & 0.5000 & 0.8200 & 1.000 & 0.4505 & 0.5000 & 0.8200 & 1.000 & 0.4505 & 0.5000 & 0.8200 & 1.000 & 0.4505 \\
    & \textsc{Summ.}& 0.5000 & 0.3400 & 1.000 & 0.2537 & 0.5000 & 0.3400 & 1.000 & 0.2537 & 0.5642 & 0.4231 & 0.4314 & 0.5638 \\
    \midrule
    \multirow{3}{*}{SAPLMA}
    & \textsc{QA}   & 0.5836 & 0.4058 & 0.5385 & 0.5498 & 0.4584 & 0.3467 & 1.000 & 0.2574 & 0.4790 & 0.4667 & 0.1346 & 0.4907 \\
    & \textsc{D2T}  & 0.5929 & 0.8200 & 1.000 & 0.4505 & 0.5330 & 0.8200 & 1.000 & 0.4505 & 0.6534 & 0.8704 & 0.7642 & 0.5953 \\
    & \textsc{Summ.}& 0.5670 & 0.3651 & 0.9020 & 0.4144 & 0.5592 & 0.3400 & 1.000 & 0.2537 & 0.5661 & 0.3421 & 0.5098 & 0.4880 \\
    \bottomrule
  \end{tabular}
  \begin{tablenotes}
      \item[] Results presented for layer / hyperparameter combinations performing best on the \textsc{Overall} task (Table \ref{tab:ragtruth-eval-per-task-llama2-7b}).
  \end{tablenotes}
  \end{threeparttable}
  }
  \caption{\texttt{RAGTruth} cross-task evaluation results - LLaMA-2 7B.}
  \label{tab:ragtruth-eval-cross-task-llama2-7b}
\end{table*}

\subsection{Experimental results}

We assess \texttt{ReDeEP}'s generalization capabilities in a cross-dataset scenario. As shown in Table \ref{tab:cross-dataset-eval}, \texttt{ReDeEP} requires specific hyperparameter tuning to perform effectively on different datasets. Although this issue is only slightly noticeable when evaluating on \texttt{RAGTruth} with hyperparameters optimized for \texttt{Dolly}, it becomes glaringly apparent when evaluating on \texttt{Dolly} using hyperparameters optimized for \texttt{RAGTruth}, where the AUC is equivalent to that of a random classifier and the correlation with the hallucination label is close to zero. The high recall on \texttt{Dolly} can be attributed to the low classification threshold employed by \texttt{ReDeEP} (approximately 0.15).


The evaluation results of our probes on LLaMA models and \texttt{RAGTruth} dataset are presented in Tables \ref{tab:ragtruth-eval-per-task-llama2-7b} and \ref{tab:ragtruth-eval-per-task-llama3-8b}. For completeness, we include results from additional model architectures in Appendix \ref{app:additional} (Tables \ref{tab:ragtruth-eval-phi35} and \ref{tab:ragtruth-eval-mistral7b}). The performance on the \texttt{SQuAD} dataset is summarized in Tables \ref{tab:squad-eval-llama2-7b} and \ref{tab:squad-eval-llama3-8b}, while the \texttt{Dolly} evaluation results are presented in Tables \ref{tab:dolly-eval-llama2} and \ref{tab:dolly-eval-llama3}.

Given our experimental results, we find that performance across hallucination methods is highly fragmented: different classifiers perform best depending on the dataset, model, or task, with no consistent winner across settings. In particular, there is no clear advantage of SoTA detection methods over simple linear probes. In many cases, linear classifiers trained on model activations match or even outperform more complex methods like \texttt{ReDeEP}, \texttt{SAPLMA} or SAE-based classifiers. This suggests that current approaches may be overfitting to task-specific artifacts rather than capturing generalizable signals of hallucination.

To further evaluate robustness, we have assessed all methods in a cross-task setting on \texttt{RAGTruth} (cf. Table \ref{tab:ragtruth-eval-cross-task-llama2-7b} and \ref{tab:ragtruth-eval-cross-task-llama3-8b}), as well as in the \texttt{RAGTruth} $\to$ \texttt{SQuAD} (cf. Table \ref{tab:inter-eval-inter-dataset-train-ragtruth-llama2-7b} and \ref{tab:inter-eval-inter-dataset-train-ragtruth-llama3-8b}) and \texttt{SQuAD} $\to$ \texttt{RAGTruth} cross-dataset settings (cf. Table \ref{tab:inter-eval-inter-dataset-train-squad-llama2-7b} and \ref{tab:inter-eval-inter-dataset-train-squad-llama3-8b}) specifically aimed at evaluating generalization capabilities. Across the board, performance dropped substantially, with no method demonstrating consistent transferability. Both SoTA and linear probes exhibit near-random performance when applied outside their training distribution. This supports our hypothesis that hallucination detectors are latching onto task- or dataset-specific cues. The failure to generalize even across similarly structured QA tasks underscores the limitations of using internal activations as a reliable signal for hallucination detection. 

\subsection{Are we really measuring the right thing?}

The problem of measuring general hallucinations based on model internals remains open. Spurious correlation is part of a larger issue. There has been some doubt about hallucination detectors. \cite{levinstein2024no} demonstrate how \texttt{SAPLMA} \cite{azaria2023internalstatellmknows} does not predict truth but rather another spurious phenomenon, such as \textit{Sentence is true and contains no negation.} Even minor changes to the test dataset, like negating sentences, make \texttt{SAPLMA}'s accuracy random. Contrast-Consistent Search (\texttt{CCS}) \cite{burns2024discoveringlatentknowledgelanguage} finds a direction in the activation space that satisfies logical consistency properties, such as having opposite truth values for a statement and its negation. However, these axioms are insufficient, and the \texttt{CCS} probe identifies sentences with negations \cite{levinstein2024no, farquhar2023challengesunsupervisedllmknowledge}. Another line uses uncertainty quantification metrics, such as perplexity, length-normalized entropy \cite{sun2025redeep}, semantic entropy \cite{kossen2024semanticentropyprobesrobust}, and P(true) \cite{kadavath2022languagemodelsmostlyknow}. However, uncertainty metrics correlate with sequence length, skewing evaluations \cite{santilli2024on}. The overall goal in hallucination detection is reminiscent of the controversial status of polygraphs \cite{lykken1998tremor}. There is no consensus on whether hallucination behavior in a general context can be detected using current MI methods. While simpler behaviors like refusal \cite{arditi2024refusallanguagemodelsmediated} or sentiment \cite{tigges2023linearrepresentationssentimentlarge} have been explained by a single linear direction, the hallucination phenomenon is less clearly defined.

\subsection{Evaluation guidelines}

We propose a number of best practices for future work in mechanistic interpretability of hallucinations. First, consider a rigorous mathematical definition of a hallucination. In the absence of one, it is challenging to design a detector. Then consider what follows. \textbf{(I)} Check against na\"ive classifiers based on training set features (see Section \ref{sec:naive}) and \textbf{(II)} against simple linear probes. \textbf{(III)} A detector method should be trained or tuned on a specific dataset and evaluated on a different one. This also applies to unsupervised methods, like \texttt{ReDeEP}, which depend on dataset-specific hyperparameters for optimal performance (see Appendix I in \texttt{ReDeEP} \cite{sun2025redeep}). An example of that would be training / tuning on \texttt{SQuAD} and evaluating on \texttt{RAGTruth}. \textbf{(IV)} Verify if the suspected \textit{truth circuit} satisfies logical requirements like negation and de Morgan rules. \textbf{(V)} Attempt to highlight the incorrect part of the answer. This can be done using external BERT models, as seen in \cite{kovács2025lettucedetecthallucinationdetectionframework}. The \texttt{RAGTruth} dataset has span-level labels that allow testing this. To our knowledge, there is no purely activation-based detection method evaluated in this manner.

Also worth considering is that LLMs are alike imagination engines - hallucination enables the exploration of ideas and options. In this light, detecting hallucinations could be related to notions of orthogonality between the generated answer and the context provided to the model.

We find the hurdles in hallucination detection to be similar to the challenges in evaluating adversarial defenses \cite{carlini2019evaluatingadversarialrobustness}. \cite{carlini2020adversarial} suggest that instead of trying to find a method for all cases, it is better to restrict the scope. For hallucination detection, we should shift the attention to finding specific, clearly defined sub-hallucinations. \textbf{(1)} A method to verify if a given entity is known to the model \cite{ferrando2024iknowentityknowledge}. \textbf{(2)} A mechanism by which LLMs switch between contextual and parametric knowledge \cite{zhao2024steeringknowledgeselectionbehaviours, minder2025controllablecontextsensitivityknob}.

\section{Conclusion}
We have assessed general hallucination detectors and found that SOTA performance on the \texttt{RAGTruth} dataset may be overstated due to evaluation problems. On top of that, even in more rigorous settings, SOTA is often outperformed in-distribution by  linear probes. Furthermore, each of the considered methods performs no better than random out-of-distribution. This underscores the challenge in general hallucination detection, where current methods based on model internals are unlikely to generalize to unseen data. In light of this challenge, we propose a set of guidelines for future hallucination detection methods, in particular: restricting the scope of the hallucination and focusing on specific hallucination spans rather than general labels - potentially promising angles of attack.

\section*{Limitations}
This work is limited in that out-of-distribution generalization is currently not achieved by any method we are aware of but the introduction of new detection methods might change this state of affairs, so it is not a definite statement on the impossibility of such methods. Additionally, while we have striven to provide a comprehensive assessment of cross-dataset performance, it may be possible to show that there are approaches which work significantly better than chance on specific dataset combinations, or for narrowed down specifications of what a hallucination is, e.g. only untruthful answers, like is done in the \texttt{Truthful QA} dataset \cite{lin2022truthfulqa}.

\bibliography{references}

\appendix

\section{Hallucination types}

Hallucinations are a byproduct of all stages of the capability acquisition process in LLMs. They originate from artifacts in pre-training data, the training and alignment procedures, the decoding strategies \cite{huang2025survey}. In-context learning \cite{brown2020languagemodelsfewshotlearners} was introduced as a strategy to mitigate hallucinations in LLMs by grounding generation in externally provided information at inference time. Yet, hallucinations occur even in the presence of adequate context. Recent analyses \cite{huang2025survey} suggest these failures stem not only from missing or irrelevant information, but from how the models internally manage and integrate contextual input with their parametric knowledge. These hallucinations may reflect breakdowns in \textit{contextual awareness} - where the relevant context is available but poorly attended to, lost in long inputs or overriden by conflicting parametric knowledge - and \textit{contextual alignment} - where information is misattributed or incorrectly decoded. Our work examines whether such failures can be detected from the model's internal activations.

\section{RAGTruth}
\label{app:ragtruth}
Given the ambiguity of the \textit{hallucination} concept and the potential multitude of hallucination types, the characteristics of specific datasets used for training or hyperparameter tuning of detection methods become crucial. To the best of our knowledge, the \texttt{RAGTruth} dataset \cite{niu2024ragtruth} is the only publicly-available manually-annotated RAG-based hallucination dataset. It comprises $2\,965$ prompts and $17\,790$ responses to those prompts from $6$ LLMs: \texttt{GPT-3.5-Turbo-0613}, \texttt{GPT-4-0613}, \texttt{LLaMA-2-7B-Chat}, \texttt{LLaMA-2-13B-Chat}, \texttt{LLaMA-2-70B-Chat} and \texttt{Mistral-7B-Instruct}. The data is broken down into $3$ tasks:
\begin{itemize}
    \item Question Answering (\texttt{QA}): answer questions related to daily life, pre-selected from the \texttt{MS MARCO} dataset \cite{bajaj2016msmarcohumangenerated}.
    \item Data-to-text Writing (\texttt{D2T}): provide an objective description of a randomly sampled business from the \href{https://business.yelp.com/data/resources/open-dataset/}{\texttt{Yelp Open Dataset}} where part of the prompt contains JSON-formatted data.
    \item Summarization (\texttt{Summary}): summarize a piece of news from either the \texttt{CNN/Daily Mail} dataset \cite{hermann2015teching, nallapati2016abstractive} or an undisclosed news platform.
\end{itemize}
An example of a prompt and response for the \texttt{LlaMA-2-7B-Chat} model with the hallucinated part of the response marked is visible in Table \ref{tab:ragtruth}. The responses from the models are generated for non-zero temperature and so, are not consistently reproducible. The evaluation of hallucination detection and mitigation methods on \texttt{RAGTruth} focuses on the specific rollout present in the dataset and implicitly ignores the possibility that the model may not generate hallucinatory responses for a specific prompt for the majority of sampled rollouts not present in the dataset.

In general, the distinct characteristics of the tasks (\texttt{QA}, \texttt{D2T} and \texttt{Summary}) provide a hint that each of them may contain data artifacts not present in the other ones. Additionally, hallucination incidence varies significantly between the tasks, being markedly higher for \texttt{D2T}. Those two factors combined suggest that the artifacts may correlate with the hallucination label. Consequently, the evaluation of hallucination detection methods should take this into account. At the very least, it should provide performance metrics per task rather than aggregated for the whole dataset.

\begin{table*}[t]
  \centering
  \begin{tabular}{p{0.5\textwidth} p{0.5\textwidth}}
    \toprule
    \textsc{Prompt} & \textsc{Answer} \\
    \midrule
    Instruction: \newline Write an objective overview (...) based only on the provided structured data in the JSON format. (...) \newline Structured data: \newline $\{$ \newline \hspace*{1em} 'name': 'Apna Indian Kitchen', \newline \hspace*{1em} 'address': '718 State St', \hspace*{1em} \newline \hspace*{1em} 'city': 'Santa Barbara', \newline \hspace*{1em} 'state': 'CA', \newline \hspace*{1em} (...), \newline \hspace*{1em} 'attributes': $\{$ \newline \hspace*{2em} (...), \newline \hspace*{2em} 'OutdoorSeating': True, \newline \hspace*{2em} {\color{red}'WiFi': 'no'}, \newline \hspace*{2em} (...) \newline \hspace*{1em} $\}$, \newline \hspace*{1em} 'business$\_$stars': 4.5, \newline \hspace*{1em} 'review$\_$info': $\left[\left(\dots\right)\right]$ \newline $\}$ \newline Overview: & Apna Indian Kitchen is a highly rated restaurant located in Santa Barbara, California, serving a variety of gluten-free, gastropub, vegan, and Indian cuisine. According to customer reviews, the restaurant offers tasty and authentic Indian food with great service. Customers have praised the menu options, including mixed veg spicy and butter chicken medium spicy dishes, roti, and naan. The restaurant has a cozy atmosphere with outdoor seating and {\color{red}free WiFi}. While the restrooms could use some attention, the serving sizes are perfect, and the staff is friendly. Customers have given the restaurant a 4.5-star rating, with many leaving positive reviews highlighting the quality of the food and service. \\
    \bottomrule
  \end{tabular}
  \caption{\texttt{RAGTruth D2T} example.}
  \label{tab:ragtruth}
\end{table*}

\section{SQuAD}
\label{app:squad}





Stanford Question Answering Dataset (\texttt{SQuAD}) \cite{rajpurkar2018knowdontknowunanswerable} is a reading comprehension benchmark comprising of context passages based on a set of Wikipedia articles and questions created by crowdworkers (see Appendix \ref{app:squad}, Table \ref{tab:squad} for an example). \texttt{SQuAD}'s consistent format - answering questions directly from context - helps minimize the chance of spurious task-related correlations. It resembles the QA task in \texttt{RAGTruth}, where the model answers questions based on short context excerpts but, contrary to \texttt{RAGTruth}, \texttt{SQuAD} only contains human-generated reference answers and not model generated answers reviewed by humans. In our evaluation procedure, we prompted each model to answer questions based on context. We subsequently evaluated the generated answers as hallucinatory or non-hallucinatory by comparing them to reference answers using \texttt{LLaMA-3 70B} \cite{grattafiori2024llama3herdmodels} as a judge.

\begin{table*}[t]
  \centering
  \renewcommand{\arraystretch}{1.3} 
    \begin{tabular}{p{0.5\textwidth} p{0.25\textwidth} p{0.2\textwidth}}
      \toprule
      \textsc{Context} & \textsc{Question} & \textsc{Answer} \\
      \midrule
      \justifying In many societies, beer is the most popular alcoholic drink. Various social traditions and activities are associated with beer drinking, such as playing cards, darts, or other pub games; attending beer festivals; engaging in zythology (the study of beer); visiting a series of pubs in one evening; visiting breweries; beer-oriented tourism; or rating beer. Drinking games, such as beer pong, are also popular. A relatively new profession is that of the beer sommelier, who informs restaurant patrons about beers and food pairings.  
      & What is a popular drinking game where beer is often considered?  
      & Beer pong \\
      \bottomrule
    \end{tabular}
  \caption{\texttt{SQuAD} dataset example.}
  \label{tab:squad}
\end{table*}

\section{Activation probes}
\label{app:activation-probes}
Linear probes have gained traction in the MI community as, under the \textit{linear representation} hypothesis, they are sufficient to detect features in a model's representations \cite{bereska2024mechanisticinterpretabilityaisafety}. We train a separate Logistic Regression classifier for each layer's pre-attention and pre-MLP activations, using the last token's features to predict hallucinations across the model. 

Given the high class imbalance (the hallucination rate in responses is approximately 10-20\% depending on model and dataset) and a line of work in MI hinting at existence of non-linear features in LLMs \cite{engels2025languagemodelfeaturesonedimensionally}, we decided to train a non-linear probe, a Random Forest classifier, on model activations alongside the linear probe.

\subsection{SAE probes}
\label{app:sae-probes}
SAEs have garnered considerable attention in the MI community thanks to their ability to find at least some human-interpretable features \cite{templeton2024scaling, gao2024scalingevaluatingsparseautoencoders}. It has been shown that SAE features may also store some meta information on the activity of the base model. For instance, they can, to an extent, regulate the strength of attending to the context vs. relying on parametric knowledge in LLMs \cite{zhao2024steeringknowledgeselectionbehaviours}. Similarly, SAE activations carry information about the uncertainty of the LLM about an entity it is asked about \cite{ferrando2024iknowentityknowledge}. We extract SAE activations using the SAEs provided by \cite{zhao2024steeringknowledgeselectionbehaviours} for layers $12$, $13$, $14$ and $15$ of the \texttt{LLaMA-2 7B} model. All activations are rescaled per feature to the $\left[0; 1\right]$ range. Two extraction methods are used: activations at the last token of the response and the maximum activations over the prompt and the response. Additionally, the activations are provided to the classifiers in two flavors. In the first approach, they are directly treated as input. In the second one, a contrastive representation is calculated:
\begin{equation}
   \hat{\mathbf{a}} = \mathbf{a}_{H} - \mathbf{a}_{C} 
\end{equation}
where $\hat{\mathbf{a}} \in \mathbb{R}^{d}$, $\mathbf{a}_{H}$ are the SAE activations for hallucinatory samples and $\mathbf{a}_{C}$ are the corresponding activations for non-hallucinatory samples. The dimensionality of SAE dictionary elements is $d$. We choose the top $k$ elements from $\hat{\mathbf{a}}$ with the highest magnitude on the train set. In our experiments, we set $k = 4\,096$ to match the dimensionality of the raw representation from the LLM in order to make the results more comparable with classification based on raw activations.

\section{Additional experimental results}
\label{app:additional}

We present additional experimental results in Tables \ref{tab:hallucination_all_models}-\ref{tab:inter-eval-inter-dataset-train-squad-llama3-8b}.

\begin{table*}[ht]
  \centering
  \resizebox{0.8\columnwidth}{!}{
  \begin{tabular}{lcc}
    \toprule
    \textsc{Model} & \textsc{Task Type} & \textsc{Hallucination Rate} \\
    \midrule
    GPT-3.5 Turbo (0613)  & D2T      & 0.2633 \\
                         & QA       & 0.0758 \\
                         & Summary  & 0.0573 \\
    GPT-4 (0613)         & D2T      & 0.2807 \\
                         & QA       & 0.0425 \\
                         & Summary  & 0.0785 \\
    LLaMA-2 13B Chat      & D2T      & 0.9516 \\
                         & QA       & 0.4034 \\
                         & Summary  & 0.3128 \\
    LLaMA-2 70B Chat      & D2T      & 0.8354 \\
                         & QA       & 0.3236 \\
                         & Summary  & 0.2248 \\
    LLaMA-2 7B Chat       & D2T      & 0.8596 \\
                         & QA       & 0.5157 \\
                         & Summary  & 0.4602 \\
    LLaMA-3 8B Instruct   & D2T      & 0.8800 \\
                         & QA       & 0.5200 \\
                         & Summary  & 0.2200 \\
    Mistral 7B Instruct  & D2T      & 0.9274 \\
                         & QA       & 0.3822 \\
                         & Summary  & 0.6543 \\
    \bottomrule
  \end{tabular}
  }
  \caption{Hallucination rates on \texttt{RAGTruth} per model and task type.}
  \label{tab:hallucination_all_models}
\end{table*}

\begin{table*}[ht]
  \centering
  \resizebox{0.6\columnwidth}{!}{
  \begin{tabular}{lcc}
    \toprule
    \textsc{Model} & \textsc{AUC} & \textsc{PCC} \\
    \midrule
    GPT-4 (0613)         & 0.7757 & 0.3403 \\
    GPT-3.5 Turbo (0613) & 0.7623 & 0.3372 \\
    Mistral 7B Instruct  & 0.7267 & 0.4777 \\
    LLaMA-2 7B Chat       & 0.7119 & 0.4494 \\
    LLaMA-2 13B Chat      & 0.8086 & 0.6526 \\
    LLaMA-2 70B Chat      & 0.7594 & 0.5342 \\
    LLaMA-3 8B Instruct   & 0.7084 & 0.4437 \\
    \bottomrule
  \end{tabular}
  }
  \caption{The AUC and PCC scores on naïve classifier across models on \texttt{RAGTruth}.}
  \label{tab:naive_all_models}
\end{table*}

\begin{table*}[ht]
  \centering
  \resizebox{.6\columnwidth}{!}{
  \begin{tabular}{lcc}
    \toprule
    \textsc{Classifier} & \textsc{AUC} & \textsc{PCC} \\
    \midrule
    na\"ive & 0.8086 & \textbf{0.6526} \\
    SEP & 0.8089 & 0.5276 \\
    SAPLMA & 0.8029 & 0.3956 \\
    ReDeEP (token) & 0.8181 & 0.5478 \\
    ReDeEP (chunk) & \textbf{0.8244} & 0.5566 \\
    \bottomrule
  \end{tabular}
  }
  \caption{Model performance metrics on \texttt{RAGTruth} using LLaMA-2 13B.}
  \label{tab:model_performance_13b}
\end{table*}

\begin{table*}[ht]
  \centering
  \resizebox{.6\columnwidth}{!}{
  \begin{tabular}{lcc}
    \toprule
    \textsc{Classifier} & \textsc{AUC} & \textsc{PCC} \\
    \midrule
    na\"ive & 0.7281 & \textbf{0.4824} \\
    SEP & 0.7004 & 0.3713 \\
    SAPLMA & 0.7092 & 0.4054 \\
    ReDeEP (token) & \textbf{0.7522} & 0.4493 \\
    ReDeEP (chunk) & 0.7285 & 0.3964 \\
    \bottomrule
  \end{tabular}
  }
  \caption{Model performance metrics on \texttt{RAGTruth} using LLaMA-3 8B.}
  \label{tab:model_performance_8b}
\end{table*}

\begin{table*}[ht]
  \centering
  \resizebox{\textwidth}{!}{
  \begin{threeparttable}
  \begin{tabular}{l|cccccccccccc|cccc}
    \toprule
    & \multicolumn{4}{c}{\textsc{Question Answering}} 
    & \multicolumn{4}{c}{\textsc{Data-to-text writing}} 
    & \multicolumn{4}{c}{\textsc{Summarization}} 
    & \multicolumn{4}{c}{\textsc{Overall}} \\
    \cmidrule(r){2-5} \cmidrule(r){6-9} \cmidrule(r){10-13} \cmidrule(r){14-17}
    Method & AUC & Precision & Recall & F1 
           & AUC & Precision & Recall & F1 
           & AUC & Precision & Recall & F1 
           & AUC & Precision & Recall & F1 \\
    \midrule
    ReDeEP & 0.5974 & 0.5514 & \textbf{0.7564} & 0.6378 
           & \textbf{0.5084} & \textbf{0.8707} & \textbf{0.7652} & \textbf{0.8145} 
           & 0.5593 & 0.2475 & \textbf{0.8065} & 0.3788 
           & \textbf{0.7534} & 0.6465 & \textbf{0.8000} & \textbf{0.7151} \\
    Logistic Regression\tnote{a} & \textbf{0.8750} & \textbf{0.8928} & \textbf{0.8750} & \textbf{0.8685} 
                                 & 0.5000 & 0.4347 & 0.5000 & 0.4651 
                                 & 0.5166 & 0.5197 & 0.5166 & 0.5165 
                                 & 0.7001 & 0.7045 & 0.7000 & 0.6984 \\
    Random Forest\tnote{b} & 0.8257 & 0.8257 & 0.8257 & 0.8257 
                           & 0.5000 & 0.4347 & 0.5000 & 0.4651 
                           & \textbf{0.6410} & \textbf{0.6598} & 0.6410 & \textbf{0.6376} 
                           & 0.8221 & \textbf{0.8221} & 0.8221 & 0.8221 \\
    \bottomrule
  \end{tabular}
  \begin{tablenotes}
  \item[a] Best result - based on layer $14$'s activations.
  \item[b] Best result - based on layer $8$'s activations.
  \end{tablenotes}
  \end{threeparttable}
  }
  \caption{\texttt{RAGTruth} evaluation results - LLaMA-3 8B.}
  \label{tab:ragtruth-eval-per-task-llama3-8b}
\end{table*}

\begin{table*}[ht]
  \centering
  \resizebox{\textwidth}{!}{
  \begin{threeparttable}
  \begin{tabular}{l|cccc|cccc|cccc|cccc}
    \toprule
    & \multicolumn{4}{c|}{\textsc{Question Answering}} 
    & \multicolumn{4}{c|}{\textsc{Data-to-text writing}} 
    & \multicolumn{4}{c|}{\textsc{Summarization}} 
    & \multicolumn{4}{c}{\textsc{Overall}} \\
    \cmidrule(r){2-5} \cmidrule(r){6-9} \cmidrule(r){10-13} \cmidrule(r){14-17}
    Method & AUC & Precision & Recall & F1 
           & AUC & Precision & Recall & F1 
           & AUC & Precision & Recall & F1 
           & AUC & Precision & Recall & F1 \\
    \midrule
    SAPLMA & \textbf{0.7161} & 0.5405 & 0.5000 & 0.6768 
           & 0.5277 & 0.2778 & 0.2632 & 0.5157 
           & \textbf{0.8103} & 0.6000 & 0.3750 & 0.7187 
           & \textbf{0.7303} & 0.4231 & 0.3837 & 0.6346 \\
    Logistic Regression\tnote{a} & 0.6863 & \textbf{0.7544} & \textbf{0.6920} & \textbf{0.7011} 
                        & \textbf{0.5546} & 0.5571 & \textbf{0.5571} & \textbf{0.5571} 
                        & 0.6428 & \textbf{0.7829} & \textbf{0.6448} & \textbf{0.7136} 
                        & 0.6483 & 0.6789 & \textbf{0.6401} & \textbf{0.6513} \\
    Random Forest \tnote{b}      & 0.5000 & 0.5000 & 0.5000 & 0.4100 
                        & 0.5111 & \textbf{0.8642} & 0.5125 & 0.5441 
                        & 0.5714 & 0.7428 & 0.5717 & 0.6108 
                        & 0.5097 & \textbf{0.8861} & 0.5092 & 0.4543 \\
    \bottomrule
  \end{tabular}
  \begin{tablenotes}
  \item[a] Best result - based on layer $18$'s activations.
  \item[b] Best result - based on layer $22$'s activations.
  \end{tablenotes}
  \end{threeparttable}
  }
  \caption{\texttt{RAGTruth} evaluation results - Phi3.5 Mini.}
  \label{tab:ragtruth-eval-phi35}
\end{table*}

\begin{table*}[ht]
  \centering
  \resizebox{\textwidth}{!}{
  \begin{threeparttable}
  \begin{tabular}{l|cccc|cccc|cccc|cccc}
    \toprule
    & \multicolumn{4}{c|}{\textsc{Question Answering}} 
    & \multicolumn{4}{c|}{\textsc{Data-to-text writing}} 
    & \multicolumn{4}{c|}{\textsc{Summarization}} 
    & \multicolumn{4}{c}{\textsc{Overall}} \\
    \cmidrule(r){2-5} \cmidrule(r){6-9} \cmidrule(r){10-13} \cmidrule(r){14-17}
    Method & AUC & Precision & Recall & F1 
           & AUC & Precision & Recall & F1 
           & AUC & Precision & Recall & F1 
           & AUC & Precision & Recall & F1 \\
    \midrule
    SAPLMA & \textbf{0.8905} & 0.6897 & 0.6452 & \textbf{0.7917} 
           & 0.5823 & \textbf{0.8933} & \textbf{1.0000} & 0.4718 
           & \textbf{0.6957} & 0.6356 & \textbf{0.8721} & 0.5864 
           & \textbf{0.8746} & 0.7710 & \textbf{0.9124} & \textbf{0.7904} \\
    Logistic Regression\tnote{a} & 0.6451 & 0.6400 & 0.6300 & 0.6400 
                        & 0.6120 & 0.6120 & 0.6940 & \textbf{0.7010} 
                        & 0.6291 & \textbf{0.6292} & 0.6291 & \textbf{0.6292} 
                        & 0.7818 & 0.8060 & 0.7817 & 0.7862 \\
    Random Forest\tnote{b}       & 0.6224 & \textbf{0.6200} & \textbf{0.6200} & 0.6100 
                        & \textbf{0.5439} & 0.5439 & 0.8917 & 0.5207 
                        & 0.6371 & 0.6371 & 0.6498 & \textbf{0.6245} 
                        & 0.7429 & \textbf{0.8064} & 0.7429 & 0.7451 \\
    \bottomrule
  \end{tabular}
  \begin{tablenotes}
  \item[a] Best result - based on layer $17$'s activations.
  \item[b] Best result - based on layer $24$'s activations.
  \end{tablenotes}
  \end{threeparttable}
  }
  \caption{\texttt{RAGTruth} evaluation results - Mistral 7B.}
  \label{tab:ragtruth-eval-mistral7b}
\end{table*}

\begin{table*}[t]
  \centering
  \resizebox{\textwidth}{!}{
  \begin{threeparttable}
  \begin{tabular}{l|c|cccccccccccc}
    \toprule
    & & \multicolumn{4}{c}{\textsc{Question Answering}} & \multicolumn{4}{c}{\textsc{Data-to-text writing}} & \multicolumn{4}{c}{\textsc{Summarization}} \\
    \cmidrule(r){3-6} \cmidrule(r){7-10} \cmidrule(r){11-14} 
    Method & Eval task & AUC & Precision & Recall & F1 & AUC & Precision & Recall & F1 & AUC & Precision & Recall & F1 \\
    \midrule
    \multirow{3}{*}{Logistic Regression} 
    & \textsc{QA}   & 0.6477 & 0.6500 & 0.6500 & 0.6500 & 0.5117 & 0.5300 & 0.5300 & 0.4500 & 0.5104 & 0.5400 & 0.5100 & 0.2300 \\
    & \textsc{D2T}  & 0.5479 & 0.7000 & 0.5500 & 0.5600 & 0.5000 & 0.4300 & 0.5000 & 0.4700 & 0.5878 & 0.5600 & 0.5900 & 0.5000 \\
    & \textsc{Summ.}& 0.5104 & 0.5400 & 0.5100 & 0.2300 & 0.5113 & 0.5600 & 0.5100 & 0.1300 & 0.4666 & 0.3200 & 0.4700 & 0.3600 \\
    \midrule
    \multirow{3}{*}{Random Forest} 
    & \textsc{QA}   & 0.7386 & 0.7400 & 0.7400 & 0.7400 & 0.6338 & 0.5600 & 0.6300 & 0.5500 & 0.6146 & 0.5800 & 0.6100 & 0.5800 \\
    & \textsc{D2T}  & 0.4903 & 0.4900 & 0.4900 & 0.4500 & 0.5000 & 0.4300 & 0.5000 & 0.4700 & 0.5392 & 0.5300 & 0.5400 & 0.5100 \\
    & \textsc{Summ.}& 0.5331 & 0.5600 & 0.5300 & 0.4700 & 0.4861 & 0.4800 & 0.4900 & 0.1800 & 0.5000 & 0.3300 & 0.5000 & 0.3900 \\
    \bottomrule
  \end{tabular}
  \begin{tablenotes}
      \item[] Results presented for layer / hyperparameter combinations performing best on the \textsc{Overall} task (Table \ref{tab:ragtruth-eval-per-task-llama3-8b}).
  \end{tablenotes}
  \end{threeparttable}
  }
  \caption{\texttt{RAGTruth} cross-task evaluation results - LLaMA-3 8B.}
  \label{tab:ragtruth-eval-cross-task-llama3-8b}
\end{table*}

\begin{table*}[ht]
  \centering
  \resizebox{\columnwidth}{!}{
  \begin{tabular}{lcccc}
    \toprule
    \textsc{Method} & \textsc{AUC} & \textsc{Precision} & \textsc{Recall} & \textsc{F1} \\
    \midrule
    ReDeEP & 0.5163 & 0.4946 & \textbf{0.9716} & 0.6555 \\
    Logistic Regression & \textbf{0.6862} & 0.6872 & 0.6859 & \textbf{0.6862} \\
    Random Forest & 0.6581 & 0.6656 & 0.6581 & 0.6541 \\
    SAE Classifier & 0.6684 & \textbf{0.7609} & 0.4895 & 0.6578 \\
    \bottomrule
  \end{tabular}
  } 
  \caption{\texttt{SQuAD} evaluation results – LLaMA-2 7B.}
  \label{tab:squad-eval-llama2-7b}
\end{table*}

\begin{table*}[ht]
  \centering
  \resizebox{\columnwidth}{!}{
  \begin{tabular}{lcccc}
    \toprule
    \textsc{Method} & \textsc{AUC} & \textsc{Precision} & \textsc{Recall} & \textsc{F1} \\
    \midrule
    ReDeEP & 0.5851 & 0.5330 & 0.8821 & 0.6645 \\
    Logistic Regression & 0.7000 & 0.7045 & 0.7000 & 0.6984 \\
    Random Forest & 0.6687 & 0.6687 & 0.6687 & 0.6687 \\
    \bottomrule
  \end{tabular}
  }
  \caption{\texttt{SQuAD} evaluation results – LLaMA-3 8B.}
  \label{tab:squad-eval-llama3-8b}
\end{table*}

\begin{table*}[ht]
  \centering
  \resizebox{\columnwidth}{!}{
  \begin{threeparttable}
  \begin{tabular}{lcccc}
    \toprule
    \textsc{Method} & \textsc{AUC} & \textsc{Precision} & \textsc{Recall} & \textsc{F1} \\
    \midrule
    ReDeEP & 0.5741 & 0.5714 & 0.8889 & 0.6957 \\
    Logistic Regression & 0.8055 & 0.7946 & 0.8055 & 0.7963 \\
    Random Forest & 0.8055 & 0.7946 & 0.8055 & 0.7963 \\
    \bottomrule
  \end{tabular}
  \begin{tablenotes}
  \item[] The results presented are based on a train/test split of the \texttt{Dolly} dataset, which is different that in \cite{sun2025redeep} where the whole dataset is used as the test set.
  \end{tablenotes}
  \end{threeparttable}
  }
  \caption{\texttt{Dolly} evaluation results – LLaMA-2 7B.}
  \label{tab:dolly-eval-llama2}
\end{table*}

\begin{table*}[ht]
  \centering
  \resizebox{\columnwidth}{!}{
  \begin{tabular}{lcccc}
    \toprule
    \textsc{Method} & \textsc{AUC} & \textsc{Precision} & \textsc{Recall} & \textsc{F1} \\
    \midrule
    ReDeEP & 0.6852 & 0.5000 & 1.0000 & 0.6667 \\
    Logistic Regression & 0.6900 & 0.7386 & 0.6944 & 0.7000 \\
    Random Forest & 0.7500 & 0.8750 & 0.7500 & 0.7619 \\
    \bottomrule
  \end{tabular}
  }
  \caption{\texttt{Dolly} evaluation results – LLaMA-3 8B.}
  \label{tab:dolly-eval-llama3}
\end{table*}

\begin{table*}[ht]
  \centering
  \resizebox{\textwidth}{!}{
  \begin{tabular}{l|cccccccccccc|cccc}
    \toprule
    & \multicolumn{4}{c}{\textsc{Question Answering}} & \multicolumn{4}{c}{\textsc{Data-to-text writing}} & \multicolumn{4}{c}{\textsc{Summarization}} & \multicolumn{4}{c}{\textsc{Overall}} \\
    \cmidrule(r){2-5} \cmidrule(r){6-9} \cmidrule(r){10-13} \cmidrule(r){14-17}
    Method & AUC & Precision & Recall & F1 & AUC &Precision & Recall & F1 & AUC & Precision & Recall & F1 & AUC & Precision & Recall & F1 \\
    \midrule
    ReDeEP & - & - & - & - & - & - & - & - & - & - & - & - &  0.5851 & 0.5385 & 0.7642 & 0.6318 \\
    Logistic Regression & 0.5116 & 0.5000 & 0.3200 & 0.3900 & 0.5026 & 0.5000 & 0.4300 & 0.4600 & 0.4959 & 0.5000 & 0.0 & 0.0 & 0.5038 & 0.5000 & 0.2600 & 0.3400 \\
    Random Forest & 0.4969 & 0.5000 & 0.1500 & 0.2300 & 0.5246 & 0.5000 & 0.2500 & 0.3300 & 0.5009 & 0.5000 & 0.1900 & 0.2700 & 0.4978 & 0.5000 & 0.2500 & 0.3300 \\
    \bottomrule
  \end{tabular}
  }
  \caption{Evaluation results | train: \texttt{RAGTruth} | eval: \texttt{SQuAD} - LLaMA-3 8B.}
  \label{tab:inter-eval-inter-dataset-train-ragtruth-llama3-8b}
\end{table*}

\begin{table*}[ht]
  \centering
  \resizebox{\textwidth}{!}{
  \begin{tabular}{l|cccccccccccc|cccc}
    \toprule
    & \multicolumn{4}{c}{\textsc{Question Answering}} 
    & \multicolumn{4}{c}{\textsc{Data-to-text writing}} 
    & \multicolumn{4}{c}{\textsc{Summarization}} 
    & \multicolumn{4}{c}{\textsc{Overall}} \\
    \cmidrule(r){2-5} \cmidrule(r){6-9} \cmidrule(r){10-13} \cmidrule(r){14-17}
    Method & AUC & Precision & Recall & F1 
           & AUC & Precision & Recall & F1 
           & AUC & Precision & Recall & F1 
           & AUC & Precision & Recall & F1 \\
    \midrule
    Logistic Regression & 0.5011 & 0.5000 & 0.4600 & 0.4900 
                        & 0.5126 & 0.5000 & 0.4200 & 0.4600 
                        & 0.4994 & 0.5000 & 0.5100 & 0.3900 
                        & 0.5097 & 0.5100 & 0.5097 & 0.5098 \\
    Random Forest       & 0.4820 & 0.5000 & 0.2200 & 0.4100 
                        & 0.5294 & 0.5000 & 0.4200 & 0.4400 
                        & 0.5022 & 0.5000 & 0.2600 & 0.3500 
                        & 0.5120 & 0.5091 & 0.5120 & 0.5063 \\
    \bottomrule
  \end{tabular}
  }
  \caption{Evaluation results | train: \texttt{RAGTruth} | eval: \texttt{SQuAD} - Phi3.5 Mini.}
  \label{tab:inter-eval-inter-dataset-train-ragtruth-phi35mini}
\end{table*}

\begin{table*}[ht]
  \centering
  \resizebox{\columnwidth}{!}{
  \begin{threeparttable}
  \begin{tabular}{l|c|cccccccccccc}
    \toprule
    Method & Eval task & AUC & Precision & Recall & F1 \\
    \midrule
    \multirow{4}{*}{Logistic Regression} & \textsc{QA} & 0.5061 & 0.5100 & 0.5100 & 0.5000 \\
    & \textsc{D2T} & 0.5410 & 0.5600 & 0.5100 & 0.1600 \\
    & \textsc{Summ.} & 0.4916 & 0.4900 & 0.4900 & 0.4800 \\
    & \textsc{Overall} & 0.4475 & 0.4300 & 0.4500 & 0.3700 \\
    \midrule
    \multirow{4}{*}{Random Forest} & \textsc{QA} & 0.5138 & 0.5200 & 0.5100 & 0.4900 \\
    & \textsc{D2T} & 0.5027 & 0.5300 & 0.5000 & 0.1400 \\
    & \textsc{Summ.} & 0.4990 & 0.4800 & 0.5000 & 0.3600 \\
    & \textsc{Overall} & 0.4838 & 0.5200 & 0.5300 & 0.4800 \\
    \midrule
    \multirow{4}{*}{SAE Classifier} & \textsc{QA} & 0.5000 & 0.3467 & 1.0000 & 0.2574 \\
    & \textsc{D2T} & 0.5000 & 0.8200 & 1.0000 & 0.4505 \\
    & \textsc{Summ.} & 0.5000 & 0.3400 & 1.0000 & 0.2537 \\
    & \textsc{Overall} & 0.5000 & 0.3467 & 1.0000 & 0.2574 \\
    \bottomrule
  \end{tabular}
  \begin{tablenotes}
      \item[] Results presented for layer / hyperparameter combinations performing best on \textsc{SQuAD}. For SAEs, only activations for the last token of the response were considered.
  \end{tablenotes}
  \end{threeparttable}
  }
  \caption{Evaluation results | train: \texttt{SQuAD} | eval: \texttt{RAGTruth} - LLaMA-2 7B.}
  \label{tab:inter-eval-inter-dataset-train-squad-llama2-7b}
\end{table*}

\begin{table*}[ht]
  \centering
  \resizebox{\columnwidth}{!}{
  \begin{tabular}{l|c|cccccccccccc}
    \toprule
    Method & Eval task & AUC & Precision & Recall & F1 \\
    \midrule
    \multirow{4}{*}{Logistic Regression} & \textsc{QA} & 0.5053 & 0.5100 & 0.5100 & 0.4400 \\
    & \textsc{D2T} & 0.5000 & 0.4400 & 
    0.5000 & 0.4700 \\
    & \textsc{Summ.} & 0.5589 & 0.6200 & 0.5600 & 0.3000 \\
    & \textsc{Overall} & 0.5373 & 0.6300 & 0.5400 & 0.4500 \\
    \midrule
    \multirow{4}{*}{Random Forest} & \textsc{QA} & 0.4887 & 0.4400 & 0.4900 & 0.3700 \\
    & \textsc{D2T} & 0.4924 & 0.4400 & 0.4900 & 0.4600 \\
    & \textsc{Summ.} & 0.5128 & 0.6100 & 0.5100 & 0.2100 \\
    & \textsc{Overall} & 0.5254 & 0.5700 & 0.5300 & 0.4500 \\
    \bottomrule
  \end{tabular}
  }
  \caption{Evaluation results | train: \texttt{SQuAD} | eval: \texttt{RAGTruth} - LLaMA-3 8B.}
  \label{tab:inter-eval-inter-dataset-train-squad-llama3-8b}
\end{table*}

\end{document}